\documentclass[sigconf]{acmart}
\AtBeginDocument{%
  }

\setcopyright{none}
\renewcommand\footnotetextcopyrightpermission[1]{}

\usepackage{xcolor}
\usepackage{url}
\usepackage{arydshln}
\usepackage{xspace}
\usepackage{multirow} 
\usepackage{float}
\usepackage{adjustbox}
\usepackage{tcolorbox}  
\usepackage[table]{xcolor}
\usepackage{graphicx}
\usepackage{subcaption}  
\usepackage{rotating}
\usepackage{balance}
\usepackage{enumitem}

\usepackage{algorithm}
\usepackage{algpseudocode}
\algrenewcommand\algorithmicrequire{\textbf{Input:}}
\algrenewcommand\algorithmicensure{\textbf{Output:}}

\begin{document}

\title{CalibraRL: Confidence-Calibrated Reinforcement Learning for Mitigating Hallucinations in RAG}


\author{Qianchi Zhang}
\authornote{This work was supported by Beijing Advanced Innovation Center for Future Blockchain and Privacy Computing.}
\affiliation{%
  \institution{School of Artificial Intelligence, Beihang University}
  \city{Beijing}
  \country{China}
}
\email{zhangqianchi@buaa.edu.cn}

\author{Hainan Zhang}
\authornote{Corresponding author.}
\affiliation{%
  \institution{School of Artificial Intelligence, Beihang University}
  \city{Beijing}
  \country{China}}
\email{zhanghainan@buaa.edu.cn}

\author{Liang Pang}
\affiliation{
  \institution{Institute of Computing Technology, Chinese Academy of Sciences}
  \city{Beijing}
  \country{China}
}
\email{pangliang@ict.ac.cn}

\author{Hongwei Zheng}
\affiliation{
 \institution{Beijing Academy of Blockchain and Edge Computing}
 \city{Beijing}
 \country{China}}

\author{Zhiming Zheng}
\affiliation{%
  \institution{School of Artificial Intelligence, Beihang University}
  \city{Beijing}
  \country{China}}

\begin{abstract}
Large language models (LLMs) often produce plausible but incorrect answers, motivating reinforcement learning methods that reward correct answers, penalize incorrect answers, and allow abstention. However, fixed outcome-based rewards do not distinguish incorrect responses with different confidence levels or account for differences in answering reliability across queries. Consequently, such rewards may inadequately penalize overconfident errors and fail to discourage unnecessary abstention. We propose \textbf{CalibraRL}, a reinforcement learning framework that incorporates response confidence and query-level answering evidence into reward design. For non-abstaining responses, CalibraRL combines correctness-based utility with a Brier-based penalty that encourages agreement between confidence and correctness. For abstentions, it estimates conditional answering reliability using a Beta posterior derived from the correctness of non-abstaining responses within each sampled group. The abstention reward increases with the posterior probability that the model's accuracy on non-abstaining responses is below a predefined threshold. CalibraRL reuses training rollouts without requiring additional annotations indicating whether the model knows the answer to each question. Experiments on three QA datasets demonstrate that CalibraRL improves answer precision and reduces hallucinations compared with strong baselines. Further analysis shows that CalibraRL reduces mean expected calibration error (ECE) by 30.8\% across two backbone models and three datasets, highlighting its potential to improve the reliability of retrieval-augmented generation (RAG) in web search.
\end{abstract}

\maketitle

\section{Introduction}

Large language models (LLMs) have become increasingly capable of answering knowledge-intensive questions, yet they remain prone to producing plausible but factually incorrect responses~\cite{ji2023survey,huang2025survey}. Providing external documents can improve access to relevant information, but does not guarantee that a model will use the evidence correctly or recognize when it cannot answer reliably~\cite{joren2025sufficient,lin2025variational}. Reliable question answering therefore requires balancing accurate answering with appropriate abstention, as insufficient abstention exposes users to incorrect information, whereas excessive abstention withholds answers that the model could otherwise provide correctly with the available evidence.

\begin{figure}[t]
    \centering
    \includegraphics[width=\linewidth]{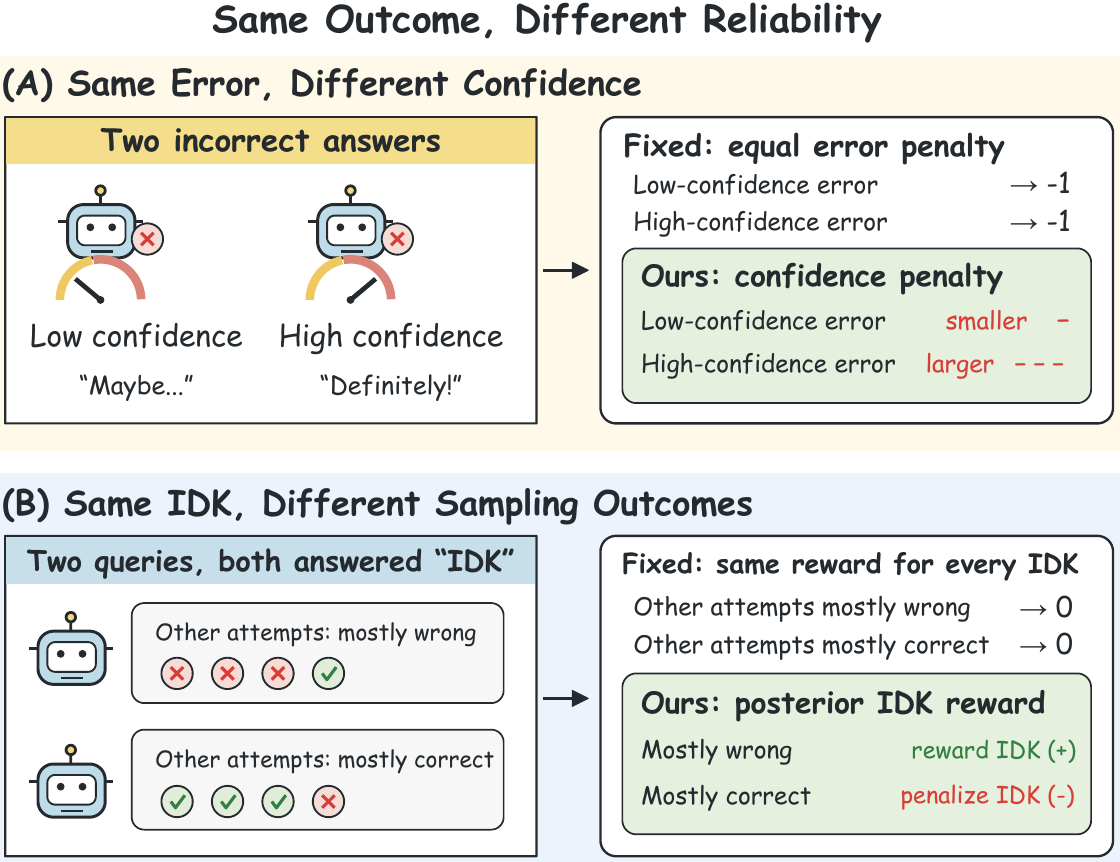}
 \caption{Reward design motivation. IDK denotes abstention. \textbf{(A)} Unlike fixed penalties, our method penalizes higher-confidence errors more strongly. \textbf{(B)} Unlike fixed abstention rewards, our posterior-adaptive reward reflects the correctness of non-abstaining rollouts for the same query.}
    \label{fig:motivation}
\end{figure}

Recent studies have explored incorporating abstention into model training~\cite{zhang2024rtuning,wei2025truthrl}. R-Tuning constructs refusal-aware instruction data to teach models to answer questions within their estimated capabilities and acknowledge uncertainty otherwise~\cite{zhang2024rtuning}. TruthRL adopts a ternary reward that rewards correct answers, penalizes incorrect answers, and assigns a neutral reward to abstentions~\cite{wei2025truthrl}. This design provides a learning signal for the trade-off between answering and abstaining. However, fixed ternary rewards assign raw reward values solely according to the final outcome category, without explicitly accounting for reliability differences within that category.
Our central observation is that the same outcome category can correspond to different reliability characteristics. As shown in Figure~\ref{fig:motivation}, two incorrect responses may have substantially different generation confidence. Although both require correction, a high-confidence error exhibits a larger discrepancy between confidence and correctness. Likewise, abstentions on different queries may be supported by different degrees of answering evidence. For one query, most of the other sampled non-abstaining responses may be incorrect, whereas for another, most may be correct. The former provides greater empirical support for abstention, while the latter suggests a potentially missed opportunity to answer correctly. These distinctions motivate training feedback that accounts not only for the observed outcome, but also for response confidence and answering evidence from the same query.

Our preliminary experiments support this motivation. First, empirical answer accuracy generally increases with generation confidence, indicating that confidence contains information about answer reliability. Nevertheless, incorrect responses span a broad confidence range, including a substantial proportion of high-confidence errors. These observations motivate examining confidence alongside correctness when designing training rewards. Second, among queries that elicit abstention, independent reference sampling reveals variation in correctness rates conditional on producing a non-abstaining response. The same abstention label therefore encompasses different empirical answering opportunities. Together, these observations motivate investigating whether confidence and answering evidence can provide useful training signals beyond outcome labels.

Motivated by these observations, we propose \textbf{CalibraRL}, a reinforcement learning framework combining confidence-aware answer rewards with posterior-adaptive abstention rewards. For non-abstaining responses, CalibraRL augments a signed correctness reward with a Brier-based penalty~\cite{glenn1950verification,damani2026beyond}. It uses generation confidence derived from token probabilities to capture disagreement between confidence and correctness, imposing a larger additional penalty on high-confidence errors. For abstentions, CalibraRL constructs a Bayesian posterior over conditional answer accuracy from the correct and incorrect non-abstaining responses in the same rollout group. The posterior probability that this accuracy falls below a specified threshold determines the abstention reward, accounting for both observed outcomes and estimation uncertainty under finite sampling. Both components are incorporated into the raw rewards used to optimize the policy through Group Relative Policy Optimization (GRPO)~\cite{shao2024deepseekmath}. The framework reuses training rollouts and reference answers for correctness verification, without requiring additional, precomputed labels of whether each query is known or unknown to the model.

Experiments on three QA datasets demonstrate that CalibraRL improves answer precision and reduces hallucinations compared with strong baselines. Further analysis shows that CalibraRL reduces mean expected calibration error (ECE) by 30.8\% across two backbone models and three datasets, highlighting its potential to improve the reliability of retrieval-augmented generation (RAG) in web search.

Our main contributions are as follows:
\begin{itemize}[nosep]
    \item We identify reliability differences within outcome categories through empirical analyses of confidence among incorrect responses and the accuracy of non-abstaining responses obtained through independent reference sampling for queries eliciting abstention.
    \item We propose CalibraRL, which combines a confidence-aware answer reward with a Bayesian posterior-adaptive abstention reward derived from same-query rollout evidence.
    \item Across three QA datasets and two backbones, we demonstrate improvements in answer precision and overall truthfulness over strong baselines, and analyze the contributions of both reward components through ablation studies.
\end{itemize}

\begin{figure*}[t!]
    \centering
    \includegraphics[width=1.0\textwidth]{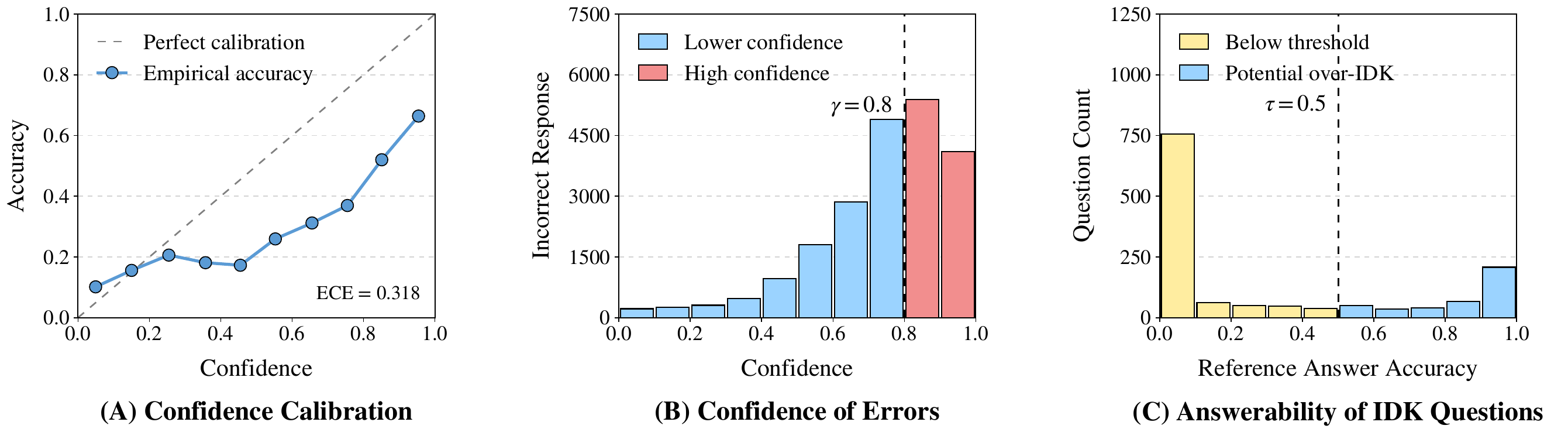}
  \caption{Empirical motivation using LLaMA3-8B-Instruct on NQ. \textbf{(A)} Empirical accuracy generally increases with confidence. \textbf{(B)} Incorrect responses exhibit heterogeneous confidence. \textbf{(C)} Questions eliciting abstention differ in the accuracy of non-abstaining responses obtained through independent reference sampling. IDK denotes \emph{``I don't know''}, indicating abstention. Reference groups containing only abstentions are excluded from (C).}
    \label{fig:motivation1}
\end{figure*}

\section{Related Work}
\subsection{LLM Hallucinations}
\label{sec:related_hallucination}

LLMs may generate fluent but factually incorrect or contextually unsupported responses~\cite{ravichander2025halogen}. Hallucinations can involve violations of factuality or faithfulness~\cite{huang2025survey}, while evaluations on TruthfulQA show that models may reproduce common human misconceptions~\cite{truthqa}. SelfCheckGPT~\cite{manakul2023selfcheckgpt} detects inconsistencies across sampled responses, whereas semantic entropy~\cite{farquhar2024detecting} measures uncertainty over semantically distinct answers. However, consistency across responses does not guarantee factual correctness. 

Existing mitigation methods improve factual reliability through retrieval augmentation~\cite{lewis2020retrieval,zhang2024adacomp,zhou2026trustworthiness,tu2025robust,chen2026improving,jin2025searchr}, response verification~\cite{dhuliawala2024cove}, contrastive decoding~\cite{chuang2024dola}, and factuality-oriented training. Self-Alignment for Factuality~\cite{zhang2024selfalignment} derives factuality-based preferences for Direct Preference Optimization (DPO)~\cite{rafailov2023direct}, while UAlign~\cite{xue2025ualign} incorporates confidence estimates and semantic entropy into reinforcement learning. Another line of research mitigates hallucinations by enabling models to abstain when they cannot answer reliably. R-Tuning~\cite{zhang2024rtuning} constructs refusal-aware training data based on knowledge gaps, while RLKF~\cite{liang2024rlkf} derives reinforcement learning feedback through knowledge-state probing and consistency checking. TruthRL~\cite{wei2025truthrl} employs Group Relative Policy Optimization (GRPO)~\cite{shao2024deepseekmath}, assigning positive rewards to correct answers, negative rewards to hallucinations, and zero rewards to abstentions. Together, these studies highlight the importance of improving factual reliability while enabling appropriate abstention.

\subsection{Knowledge Boundary of LLMs}

Knowledge-boundary research examines what models know~\cite{petroni2019language} and whether they recognize their knowledge limits~\cite{kadavath2022know}. Prior work~\cite{kadavath2022know} distinguishes response correctness probability, $P(\mathrm{True})$, from the probability of knowing an answer, $P(\mathrm{IK})$, providing evidence of models' self-assessment capabilities. However, models often express overconfidence, and reliable confidence estimation remains challenging on difficult tasks~\cite{xiong2024can}. Models also learn previously unknown facts more slowly than knowledge-consistent examples during fine-tuning, potentially increasing hallucinations~\cite{gekhman2024finetuning}. Self-evaluation research further suggests that models may fail to reliably express knowledge they possess~\cite{zhang2024selfalignment}, while knowledge-state probing examines their ability to recognize and use internal knowledge states~\cite{liang2024rlkf,ni2025towards}.

These observations distinguish knowledge availability from reliable expression: a single incorrect response does not necessarily indicate missing knowledge. A central challenge for abstention is therefore to translate answerability estimates into reliable answering or abstention decisions without unnecessarily withholding correct answers.

\section{Preliminaries}

\begin{figure*}[t!]
    \centering
    \includegraphics[width=\textwidth]{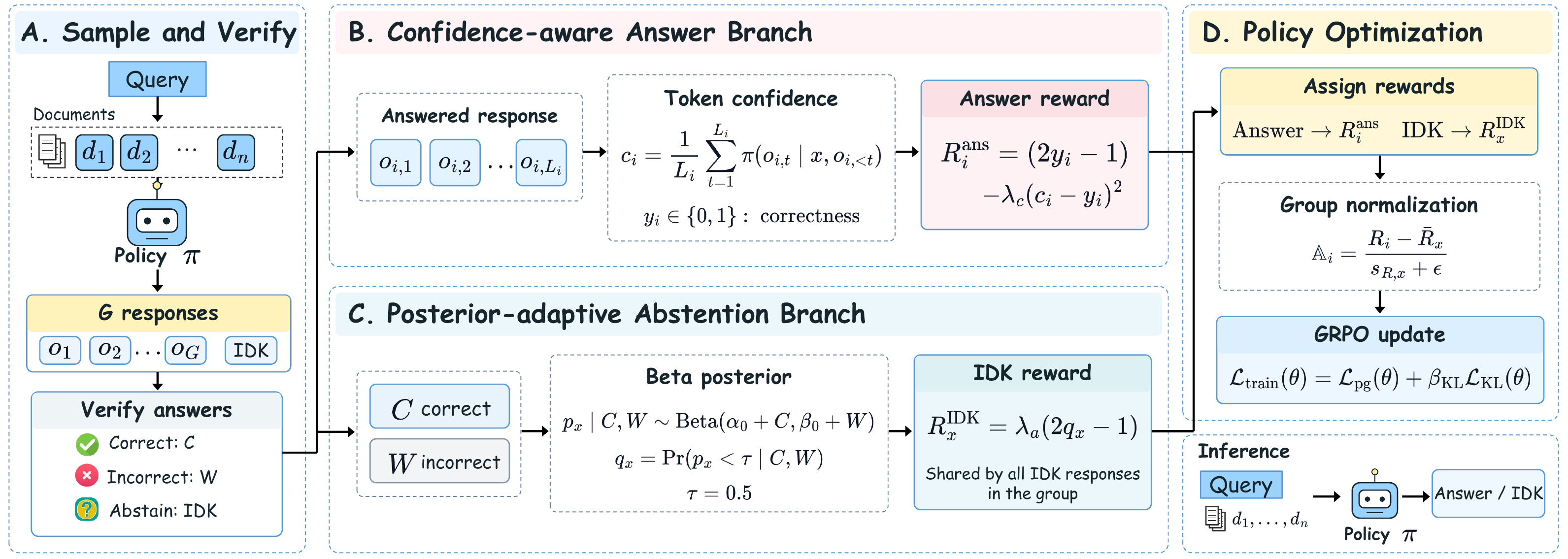}
    \caption{
    Overview of our proposed framework. IDK denotes \emph{``I don't know''}, indicating abstention.
    }
    \label{fig:framework}
\end{figure*}

\subsection{Problem Formulation}
\label{sec:problem_formulation}

We consider question answering with abstention, where a language model answers using provided documents or abstains. Let $\mathcal{D}=\{(x_j,\mathcal{A}_j)\}_{j=1}^{N}$, where $x_j=(q_j,d_j)$ contains a question and its associated documents, and $\mathcal{A}_j$ is the set of acceptable reference answers. Given $x$, an autoregressive policy $\pi_\theta$ generates $o=(o_1,\ldots,o_T)$ and is prompted to output \emph{I don't know} (IDK) when uncertain.

Let $a(o)\in\{0,1\}$ indicate abstention and $y(o,\mathcal{A})\in\{0,1\}$ denote the correctness of a non-abstaining response, assessed against $\mathcal{A}$. Responses are categorized as correct answers, incorrect answers, or abstentions. Our objective is to encourage correct answers and appropriate abstention while discouraging incorrect answers and unnecessary abstention.

Following prior work~\cite{orgad2025llms, tan-etal-2026-basecal}, we use the mean probability of generated tokens as a proxy for response confidence. For a non-abstaining response:
\begin{equation}
    c_\theta(x,o)
    =
    \frac{1}{T}
    \sum_{t=1}^{T}
    \pi_\theta(o_t\mid x,o_{<t}),
    \label{eq:response_confidence}
\end{equation}
where $T$ is the response length and $o_{<t}$ denotes preceding tokens. This measure requires no additional prompting or multiple response samples.

\subsection{Motivation}
\label{sec:motivation}

We motivate our reward design through two questions: \ding{202}~\emph{Does model confidence align with answer correctness?} \ding{203}~\emph{Does the model abstain on questions it can answer correctly?}

Figure~\ref{fig:motivation1}(A) shows that accuracy tends to increases with confidence but generally lies below the ideal calibration line, with an ECE of $0.318$. Thus, confidence is informative about correctness but exhibits substantial overconfidence. Figure~\ref{fig:motivation1}(B) shows that incorrect responses span different confidence levels, including many above $0.8$. A correctness-only reward treats these errors equally, failing to distinguish low-confidence errors from confidently incorrect responses. These observations motivate incorporating confidence--correctness alignment into the answer reward.

We next examine reference answering performance for questions that elicit IDK responses. For each question, reference accuracy is the proportion of correct responses among its non-IDK reference responses. All-IDK reference groups are excluded because this conditional accuracy is undefined. Figure~\ref{fig:motivation1}(C) shows substantial variation: many questions have low reference accuracy, but some exceed $0.5$ and others approach $1$. Thus, an IDK response does not necessarily indicate consistently unsuccessful reference answering. High reference accuracy suggests potentially unnecessary abstention, motivating an abstention reward informed by query-level answering evidence.

These findings motivate two reward-design principles: \ding{202}~\emph{penalizing discrepancies between response confidence and correctness}; \ding{203}~\emph{adapting abstention rewards to query-level answering performance}.

\section{Methodology}

As shown in Figure~\ref{fig:framework}, we sample a group of responses per query, identify abstentions, and evaluate the correctness of answered responses. For answered responses, we combine correctness utility with a confidence-based Brier penalty. For abstentions, we derive rewards from a Bayesian reliability estimate based on the group's correct and incorrect answers. Finally, the resulting rewards are normalized within each group and used for GRPO optimization.

\subsection{Confidence-Aware Answer Reward}
\label{sec:answer_reward}

For each input $x$, we sample a response group $\mathcal{G}_x=\{o_i\}_{i=1}^{G}$. For a non-abstaining response $o_i$, let $y_i=y(o_i,\mathcal{A})$ denote its correctness and $c_i$ denote its confidence computed according to \eqref{eq:response_confidence}. A correctness-only reward does not distinguish between responses with the same correctness label but different confidence levels. We therefore define the answer reward as:
\begin{equation}
    r_{\mathrm{ans}}(o_i,x)
    =
    2y_i-1-\lambda_c(c_i-y_i)^2,
    \label{eq:answer_reward}
\end{equation}
where $\lambda_c\geq 0$ controls the strength of the confidence penalty.

The correctness utility $2y_i-1$ assigns $+1$ to correct answers and $-1$ to incorrect answers. The squared-error term is based on the Brier score~\cite{glenn1950verification} and penalizes discrepancies between response confidence and correctness. Equivalently,
\begin{equation}
    r_{\mathrm{ans}}(o_i,x)
    =
    \begin{cases}
        1-\lambda_c(1-c_i)^2, & y_i=1 \\
        -1-\lambda_c c_i^2, & y_i=0
    \end{cases}.
    \label{eq:answer_reward_cases}
\end{equation}
When $\lambda_c>0$, higher confidence increases rewards for correct answers and strengthens penalties for incorrect answers. Rewards for incorrect answers never exceed $-1$, so lowering confidence cannot make these rewards positive. The penalty thus provides finer reward distinctions among responses that would otherwise receive the same correctness-based reward.

We use raw confidence scores without post-hoc calibration and treat them as fixed when computing policy gradients. These scores influence policy updates through the rewards and the resulting group-relative advantages.

\subsection{Posterior-Adaptive Abstention Reward}
\label{sec:abstention_reward}

To adapt abstention rewards to query-level answering performance, we estimate answering reliability from the non-abstaining responses in each sampled group. Let $\mathcal{I}_x=\{i:a(o_i)=0\}$ denote the index set of non-abstaining responses. The numbers of correct and incorrect answers are:
\begin{equation}
    C_x=\sum_{i\in\mathcal{I}_x}y_i,
    \qquad
    W_x=\sum_{i\in\mathcal{I}_x}(1-y_i).
    \label{eq:group_counts}
\end{equation}
Abstentions are excluded because they provide no answer whose correctness can be evaluated.

Let $p_x$ denote the probability of a correct answer conditional on answering under the policy and sampling configuration used to generate the group:
\begin{equation}
    p_x
    =
    \Pr(y=1\mid x,a(o)=0).
    \label{eq:conditional_reliability}
\end{equation}
We model the correctness labels of non-abstaining responses as conditionally independent Bernoulli observations with success probability $p_x$. With a prior $p_x\sim\operatorname{Beta}(\alpha_0,\beta_0)$, where $\alpha_0,\beta_0>0$, the posterior is:
\begin{equation}
    p_x\mid\mathcal{G}_x
    \sim
    \operatorname{Beta}(\alpha_0+C_x,\beta_0+W_x).
    \label{eq:group_posterior}
\end{equation}

For a reliability threshold $\tau\in(0,1)$, we compute the posterior probability that conditional answer accuracy falls below the threshold:
\begin{equation}
    q_x
    =
    \Pr(p_x<\tau\mid\mathcal{G}_x)
    =
    I_\tau(\alpha_0+C_x,\beta_0+W_x),
    \label{eq:posterior_tail}
\end{equation}
where $I_\tau$ is the regularized incomplete beta function. The posterior incorporates both the observed correctness outcomes and the number of non-abstaining responses available in the group.

We map this probability to an abstention reward:
\begin{equation}
    r_{\mathrm{abs}}(x,\mathcal{G}_x)
    =
    \lambda_a(2q_x-1),
    \label{eq:abstention_reward}
\end{equation}
where $\lambda_a\in[0,1]$ controls the reward magnitude. The reward lies in $[-\lambda_a,\lambda_a]$. For $\lambda_a>0$, it is positive when $q_x>1/2$ and increases with the posterior probability that answering reliability falls below $\tau$. Thus, stronger evidence of unreliable answering leads to a higher abstention reward, whereas stronger evidence of reliable answering leads to a lower reward. All abstentions within the same group share this reward.

For each sampled group, we compute the posterior from the same fixed prior and the correctness labels obtained by comparing non-abstaining responses with the ground-truth answers. If a group contains only abstentions, $C_x=W_x=0$, and the posterior remains equal to the prior. Under the default uniform prior $\alpha_0=\beta_0=1$ and threshold $\tau=0.5$, such a group receives zero abstention reward. Its identical rewards yield zero group-relative advantages and hence no reward-driven learning signal.

\subsection{Confidence-Calibrated Policy Optimization}
\label{sec:optimization_inference}

We adopt GRPO~\cite{shao2024deepseekmath} as our optimization backbone, which computes advantages from relative rewards within a response group without requiring a separate value model. For each input $x$, a behavior policy $\pi_{\theta_{\mathrm{old}}}$ samples a group $\mathcal{G}_x=\{o_i\}_{i=1}^{G}$. We combine the confidence-aware answer reward and the posterior-adaptive abstention reward to assign a scalar reward to each response:
\begin{equation}
    r_i
    =
    \begin{cases}
        r_{\mathrm{abs}}(x,\mathcal{G}_x), & a(o_i)=1, \\
        r_{\mathrm{ans}}(o_i,x), & a(o_i)=0.
    \end{cases}
    \label{eq:combined_reward}
\end{equation}
We set $\lambda_c=\lambda_a=0.5$ by default. With these weights, correct answers receive rewards no lower than abstentions, while abstentions receive higher rewards than incorrect answers. The reward gaps vary with response confidence and group-level answering reliability.

Given the rewards $\{r_i\}_{i=1}^{G}$, the group-relative advantage is:
\begin{equation}
    \mathbb{A}_i
    =
    \frac{r_i-\bar{r}}{s_r+\varepsilon_{\mathrm{std}}},
    \qquad
    \bar{r}
    =
    \frac{1}{G}\sum_{i=1}^{G}r_i,
    \label{eq:group_advantage}
\end{equation}
where $s_r$ is the within-group reward standard deviation and $\varepsilon_{\mathrm{std}}>0$ ensures numerical stability. Each response-level advantage is shared across all tokens of that response.

For the $t$-th token of response $o_i$, the importance ratio between the current and behavior policies is:
\begin{equation}
    \rho_{i,t}(\theta)
    =
    \frac{\pi_\theta(o_{i,t}\mid x,o_{i,<t})}
    {\pi_{\theta_{\mathrm{old}}}(o_{i,t}\mid x,o_{i,<t})}.
    \label{eq:importance_ratio}
\end{equation}
The clipped surrogate contribution is:
\begin{equation}
\begin{aligned}
    \ell_{i,t}(\theta)
    =
    \min\Big\{
        &\rho_{i,t}(\theta)\mathbb{A}_i,\\
        &\operatorname{clip}
        \big(\rho_{i,t}(\theta),1-\epsilon,1+\epsilon\big)
        \mathbb{A}_i
    \Big\},
\end{aligned}
    \label{eq:clipped_surrogate}
\end{equation}
where $\epsilon>0$ controls the clipping range. We compute the policy loss by averaging over tokens within each response and then over responses within each group:
\begin{equation}
    \mathcal{L}_{\mathrm{pg}}(\theta)
    =
    -\mathbb{E}_{x,\mathcal{G}_x}
    \left[
        \frac{1}{G}\sum_{i=1}^{G}
        \frac{1}{T_i}\sum_{t=1}^{T_i}
        \ell_{i,t}(\theta)
    \right],
    \label{eq:grpo_policy_loss}
\end{equation}
where $T_i$ is the response length, and the expectation is over training inputs and response groups sampled from the behavior policy.

We regularize deviations from a fixed reference policy $\pi_{\mathrm{ref}}$ using:
\begin{equation}
    \mathcal{L}_{\mathrm{KL}}(\theta)
    =
    \mathbb{D}_{\mathrm{KL}}(\pi_\theta\|\pi_{\mathrm{ref}}),
    \label{eq:grpo_kl_loss}
\end{equation}
where $\mathbb{D}_{\mathrm{KL}}$ denotes the KL regularization term between the current and reference policies. The total training loss is:
\begin{equation}
    \mathcal{L}_{\mathrm{train}}(\theta)
    =
    \mathcal{L}_{\mathrm{pg}}(\theta)
    +
    \beta_{\mathrm{KL}}\mathcal{L}_{\mathrm{KL}}(\theta),
    \label{eq:training_loss}
\end{equation}
where $\beta_{\mathrm{KL}}\geq 0$ controls the regularization strength. The confidence penalty and the adaptive abstention reward influence policy optimization through the group-relative advantages. Rewards and advantages are treated as fixed during policy updates, with no gradients propagated through the confidence scores or posterior estimates used to compute the rewards.
\subsection{Inference}
At inference time, the trained policy takes a question and its associated documents as input and directly generates a concise answer or an abstention. The prompt in Appendix~\ref{app:prompt} instructs the model to answer using the provided documents only when it is confident that its answer is correct. Otherwise, it should output \emph{``I don’t know''}. The model makes this decision during generation, without applying an explicit confidence threshold. Inference does not require ground-truth answers, group-level reliability estimation, or reward computation.

\section{Experiments}

\begin{table*}[!t]
\centering
\small
\renewcommand\arraystretch{1.0}
\caption{Main results (\%) on NQ, TriviaQA, and HotpotQA using
LLaMA3-8B-Instruct and Qwen2.5-7B-Instruct as base models.
Results are reported in terms of \textbf{\textit{Prec.}}, \textbf{\textit{Truth.}}, and \textbf{\textit{ER}},
with higher values indicating better performance for all metrics.
Avg.\ denotes the unweighted mean across the three benchmarks. The best results are in bold and the second are \underline{underlined}.}
\resizebox{\textwidth}{!}{
\begin{tabular}{ l ccc ccc ccc ccc}
\toprule
\multirow{2}[2]{*}{\textbf{Method}} 
& \multicolumn{3}{c}{\textbf{NQ}} 
& \multicolumn{3}{c}{\textbf{TriviaQA}} 
& \multicolumn{3}{c}{\textbf{HotpotQA}}
& \multicolumn{3}{c}{\textbf{Avg.}}\\

\cmidrule(r){2-4} \cmidrule(r){5-7}
\cmidrule(r){8-10} \cmidrule(r){11-13}

& \textit{Prec.}~$\uparrow$  & \textit{Truth.}~$\uparrow$  & \textit{ER}~$\uparrow$ 
& \textit{Prec.}~$\uparrow$  & \textit{Truth.}~$\uparrow$  & \textit{ER}~$\uparrow$
& \textit{Prec.}~$\uparrow$  & \textit{Truth.}~$\uparrow$  & \textit{ER}~$\uparrow$
& \textit{Prec.}~$\uparrow$  & \textit{Truth.}~$\uparrow$  & \textit{ER}~$\uparrow$  \\

\hline
\rowcolor[HTML]{F0F0F0}
\multicolumn{13}{c}{\mbox{\textit{Base Model: LLaMA3-8B-Instruct}}} \\

Prompting  & 42.88 & 63.43 & -9.11 & 68.26 & 76.22 & 27.36 & 38.76 & 68.38 & -11.61 & 49.97 & 69.34 & 2.21 \\

ICL  & 13.17 & 36.45 & -53.91 & 34.10 & 58.61 & -19.97 & 6.11 & 33.61 & -62.07 & 17.79 & 42.89 & -45.32 \\

SFT   & 43.71 & 43.88 & -12.55 & 63.98 & 63.98 & 27.96 & 34.23 & 34.43 & -31.45 & 47.31 & 47.43 & -5.35 \\

R-Tuning  & 61.36 & \underline{85.96} & 8.25 & 75.87 & \underline{84.12} & 34.05 & 51.45 & \underline{81.77} & 1.09 & 62.89 & \underline{83.95} & 14.46 \\

DTA   & 49.88 & 66.59 & -0.17 & 74.97 & 80.80 & 38.30 & 35.57 & 58.07 & -18.79 & 53.47 & 68.49 & 6.45 \\

Binary   & 49.50 & 49.53 & -1.00 & 73.54 & 73.54 & \underline{47.06} & 38.46 & 38.57 & -23.04 & 53.83 & 53.88 & 7.67 \\
TruthRL   & \underline{63.48} & 80.61 & \underline{14.32} & \underline{78.35} & 81.12 & \textbf{49.44} & \underline{51.95} & 77.30 & \underline{1.84} & \underline{64.59} & 79.68 & \textbf{21.87} \\

\cdashline{1-13}
\textbf{CalibraRL (Ours)} & \textbf{68.65} & \textbf{87.89} & \textbf{14.40} & \textbf{80.27}& \textbf{85.03} & 45.94 & \textbf{57.10} & \textbf{84.46} & \textbf{5.14} & \textbf{68.67} & \textbf{85.79} & \underline{21.83} \\

\hline

\rowcolor[HTML]{F0F0F0}

\multicolumn{13}{c}{\mbox{\textit{Base Model: Qwen2.5-7B-Instruct}}} \\

Prompting  & 42.96 & 54.24 & -11.30 & 69.36 & 74.47 & 32.27 & 42.81 & 65.48 & -8.68 & 51.71 & 64.73 & 4.10 \\

ICL  & 41.32 & 51.66 & -14.29 & 67.52 & 71.88 & 30.35 & 37.02 & 53.95 & -18.98 & 48.62 & 59.16 & -0.97\\

SFT   & 40.18 & 40.94 & -19.39 & 63.36 & 63.39 & 26.69 & 33.65 & 34.27 & -32.39 & 45.73 & 46.20 & -8.36 \\

R-Tuning  & \underline{54.72} & \textbf{77.15} & 4.76 & 74.23 & 80.54 & 36.60 & 48.81 & 74.16 & -1.20 & 59.25 & \underline{77.28} & 13.39 \\

DTA  & 45.12 & 54.82 & -8.03 & 71.32 & 75.67 & 36.17 & 41.89 & 62.59 & -10.45 & 52.78 & 64.36 & 5.90\\

Binary   & 48.06 & 48.09 & -3.88 & 74.79 & 74.79 & \textbf{49.58} & 42.96 & 42.96 & -14.07 & 55.27 & 55.28 & 10.54 \\
TruthRL   & 54.69 & 74.04 & \underline{5.37} & \underline{75.60} & \underline{80.75} & 40.40 & \underline{50.51} & \underline{76.61} & \underline{0.48} & \underline{60.27} & 77.13 & \underline{15.42} \\
\cdashline{1-13}
\textbf{CalibraRL (Ours)} & \textbf{56.49} & \underline{76.43} & \textbf{7.04}& \textbf{80.41} & \textbf{84.61} & \underline{47.77} & \textbf{58.67} & \textbf{81.48 }& \textbf{7.77} & \textbf{65.19} & \textbf{80.84} & \textbf{20.86} \\
\bottomrule
\end{tabular}
}

\label{tab:main_results}
\end{table*}

\subsection{Experimental Setup}
\subsubsection{Datasets}
We evaluate our method on three widely used QA benchmarks
covering different reasoning settings.
For open-domain QA, we adopt Natural Questions (NQ)~\cite{nq}
and TriviaQA~\cite{tqa}, two representative benchmarks
for evaluating factual question answering.
To further assess the model's ability to integrate evidence
across multiple documents, we conduct experiments on
HotpotQA~\cite{hotpotqa}, a representative multi-hop QA benchmark.
Together, these datasets enable us to evaluate our method
across factual question answering and more challenging
multi-hop reasoning scenarios.
Detailed dataset statistics are reported in Appendix~\ref{app:datasets}.


\subsubsection{Evaluation Metrics}
We evaluate reliability and confidence calibration using four metrics:
\textit{Precision} (\textbf{\textit{Prec.}})~\cite{xue2025ualign,sun2025divide},
which measures accuracy among non-abstained responses;
\textit{Truthfulness} (\textbf{\textit{Truth.}})~\cite{xu2024rejection,zhang2024selfalignment},
which measures the proportion of correct answers or abstentions;
\textit{Effective Reliability} (\textbf{\textit{ER}})~\cite{feng2024don,wen2025know},
which balances correct answering against incorrect answering;
and \textit{Expected Calibration Error} (\textbf{\textit{ECE}})~\cite{guo2017calibration},
which measures confidence calibration on non-abstained responses.
Higher values indicate better performance for Prec., Truth., and ER,
whereas lower ECE is better.
Detailed definitions and computation procedures are provided in
Appendix~\ref{app:evaluation_metrics}.

\subsubsection{Baselines}
We compare our method with the following baselines on the same test set.
Training-free methods include In-Context Learning (\textbf{ICL}),
which uses answer-and-abstention demonstrations, and
\textbf{Prompting}, which instructs the model to answer when
confident and abstain otherwise.
SFT-based methods include standard Supervised Fine-tuning
(\textbf{SFT}), which trains the model to answer every question,
and \textbf{R-Tuning}~\cite{zhang2024rtuning}, which teaches
the model to abstain beyond its knowledge. RL-based methods include
\textbf{DTA}~\cite{sun2025divide}, which models knowledge
boundaries using a four-quadrant partition and employs
DPO~\cite{rafailov2023direct};
\textbf{Binary}, which uses GRPO~\cite{shao2024deepseekmath}
with rewards of 1 for correct answers and 0 otherwise;
and \textbf{TruthRL}~\cite{wei2025truthrl}, which uses GRPO
with rewards of 1, 0, and -1 for correct answers,
abstentions, and incorrect answers, respectively.

\begin{table*}[!t]
\centering
\small
\renewcommand{\arraystretch}{1.0}
\caption{
    Ablation results (\%) on three QA benchmarks.
    \textit{w/o Answer Confidence} removes the confidence reward term;
    \textit{w/o Adaptive Abstention} replaces the adaptive abstention
    reward with a fixed reward of zero.
    Avg.\ denotes the unweighted mean across the three benchmarks.
    Best results are shown in \textbf{bold}.
}
\resizebox{\textwidth}{!}{
\begin{tabular}{l ccc ccc ccc ccc}
\toprule
\multirow{2}[2]{*}{\textbf{Variant}}
& \multicolumn{3}{c}{\textbf{NQ}}
& \multicolumn{3}{c}{\textbf{TriviaQA}}
& \multicolumn{3}{c}{\textbf{HotpotQA}}
& \multicolumn{3}{c}{\textbf{Avg.}} \\

\cmidrule(r){2-4}
\cmidrule(r){5-7}
\cmidrule(r){8-10}
\cmidrule(r){11-13}

& \textit{Prec.}~$\uparrow$
& \textit{Truth.}~$\uparrow$
& \textit{ER}~$\uparrow$
& \textit{Prec.}~$\uparrow$
& \textit{Truth.}~$\uparrow$
& \textit{ER}~$\uparrow$
& \textit{Prec.}~$\uparrow$
& \textit{Truth.}~$\uparrow$
& \textit{ER}~$\uparrow$
& \textit{Prec.}~$\uparrow$
& \textit{Truth.}~$\uparrow$
& \textit{ER}~$\uparrow$ \\

\hline
\rowcolor[HTML]{F0F0F0}
\multicolumn{13}{c}{\mbox{\textit{Base Model: LLaMA3-8B-Instruct}}} \\

\textbf{CalibraRL (Ours)}
& \textbf{68.65} & \textbf{87.89} & \textbf{14.40}
& \textbf{80.27} & \textbf{85.03} & 45.94
& \textbf{57.10} & \textbf{84.46} & \textbf{5.14}
& \textbf{68.67} & \textbf{85.79} & \textbf{21.83} \\

\textit{w/o} Answer Confidence
& 66.97 & 86.50 & 13.87
& 78.97 & 83.30 & \textbf{46.01}
& 56.05 & 83.48 & 4.55
& 67.33 & 84.43 & 21.48 \\

\textit{w/o} Adaptive Abstention
& 65.81 & 84.65 & 14.20
& 79.17 & 83.58 & 45.99
& 56.26 & 83.52 & 4.71
& 67.08 & 83.92 & 21.63\\

\hline
\rowcolor[HTML]{F0F0F0}
\multicolumn{13}{c}{\mbox{\textit{Base Model: Qwen2.5-7B-Instruct}}} \\

\textbf{CalibraRL (Ours)}
& \textbf{56.49} & \textbf{76.43} & \textbf{7.04}
& \textbf{80.41} & \textbf{84.61} & \textbf{47.77}
& \textbf{58.67} & \textbf{81.48} & 7.77
& \textbf{65.19} & \textbf{80.84} & \textbf{20.86} \\

\textit{w/o} Answer Confidence
& 56.19 & 75.32 & 6.98
& 79.17 & 83.74 & 45.54
& 58.21 & 79.76 & \textbf{7.95}
& 64.52 & 79.61 & 20.16 \\

\textit{w/o} Adaptive Abstention
& 55.48 & 74.97 & 6.16
& 79.86 & 84.12 & 47.09
& 58.18 & 80.46 & 7.64
& 64.51 & 79.85 & 20.30 \\

\bottomrule
\end{tabular}
}
\label{tab:ablation}
\end{table*}

\subsubsection{Implementation}
We use LLaMA3-8B-Instruct~\cite{grattafiori2024llama} and Qwen2.5-7B-Instruct~\cite{qwen2025qwen25technicalreport} as the backbone models following prior work~\cite{cao2025autohall,wu2026step,zhang2026redippo,xi2026confidence}. To ensure fair and consistent comparisons~\cite{cuconasu2024power,salemi2024evaluating}, we follow prior work~\cite{xu2024recomp,zhang2026stable,zhang2026less} and provide all baselines and our method with the same Top-5 Wikipedia passages retrieved by Contriever-MS MARCO~\cite{izacard2021unsupervised}. Specifically, the retrieved passages are kept fixed across different methods for each query, ensuring that all methods have access to identical external evidence. All training experiments are conducted on four NVIDIA RTX PRO 6000 GPUs. Furthermore, we use the same data splits, retrieved contexts, generation settings, and evaluation protocols across all experiments to ensure consistent and reproducible evaluations. Additional implementation details are provided in Appendix~\ref{app:training}.
\subsection{Main Results}
\label{sec:main_results}

We compare our method with all baselines on three QA
benchmarks using two backbone models.
Table~\ref{tab:main_results} summarizes the main results.

\noindent\textbf{(i) Overall performance.}
Our method achieves the highest precision in all six
model--dataset settings and the highest truthfulness
in five of them.
These results indicate higher accuracy among answered
questions and a lower proportion of incorrect outputs.

\noindent\textbf{(ii) Effectiveness on multi-hop QA.}
Our method outperforms all baselines on all three
metrics on HotpotQA with both backbone models.
The improvements in precision and ER indicate that
our method achieves higher accuracy on answered questions
while also improving the balance between correct and
incorrect answers across the full evaluation set.

\noindent\textbf{(iii) Consistency across backbone models.}
Our method achieves higher precision and truthfulness
than TruthRL on all three benchmarks with both backbone
models, demonstrating consistent gains across the two
evaluated backbones.

\begin{table*}[t]
\centering
\small
\renewcommand{\arraystretch}{1.12}
\caption{Effect of SFT initialization. CalibraRL applies reinforcement learning directly to the Instruct backbone, whereas SFT + CalibraRL first performs SFT following R-Tuning and then applies CalibraRL. All scores are reported as percentages, with higher values indicating better performance. Avg.\ denotes the unweighted mean across the three benchmarks. Best results are shown in \textbf{bold}.}
\setlength{\tabcolsep}{5pt}
\resizebox{\textwidth}{!}{
\begin{tabular}{l ccc ccc ccc ccc}
\toprule
\multirow{2}{*}{\textbf{Strategy}}
& \multicolumn{3}{c}{\textbf{NQ}}
& \multicolumn{3}{c}{\textbf{TriviaQA}}
& \multicolumn{3}{c}{\textbf{HotpotQA}}
& \multicolumn{3}{c}{\textbf{Avg.}} \\

\cmidrule(lr){2-4}
\cmidrule(lr){5-7}
\cmidrule(lr){8-10}
\cmidrule(lr){11-13}

& \textit{Prec.}~$\uparrow$
& \textit{Truth.}~$\uparrow$
& \textit{ER}~$\uparrow$
& \textit{Prec.}~$\uparrow$
& \textit{Truth.}~$\uparrow$
& \textit{ER}~$\uparrow$
& \textit{Prec.}~$\uparrow$
& \textit{Truth.}~$\uparrow$
& \textit{ER}~$\uparrow$
& \textit{Prec.}~$\uparrow$
& \textit{Truth.}~$\uparrow$
& \textit{ER}~$\uparrow$ \\
\hline

\rowcolor[HTML]{F0F0F0}
\multicolumn{13}{c}{
    \mbox{\textit{Base Model: LLaMA3-8B-Instruct}}
} \\

SFT + CalibraRL
& 58.92 & 74.68 & 11.00
& 74.15 & 77.20 & 42.59
& 47.22 & 72.41 & $-2.91$
& 60.10 & 74.76 & 16.89 \\

CalibraRL
& \textbf{68.65} & \textbf{87.89} & \textbf{14.40}
& \textbf{80.27} & \textbf{85.03} & \textbf{45.94}
& \textbf{57.10} & \textbf{84.46} & \textbf{5.14}
& \textbf{68.67} & \textbf{85.79} & \textbf{21.83} \\
\hline

\rowcolor[HTML]{F0F0F0}
\multicolumn{13}{c}{
    \mbox{\textit{Base Model: Qwen2.5-7B-Instruct}}
} \\

SFT + CalibraRL
& 50.41 & 67.87 & 0.53
& 77.74 & 81.43 & 46.27
& 56.39 & 77.45 & 6.61
& 61.51 & 75.58 & 17.80 \\

CalibraRL
& \textbf{56.49} & \textbf{76.43} & \textbf{7.04}
& \textbf{80.41} & \textbf{84.61} & \textbf{47.77}
& \textbf{58.67} & \textbf{81.48} & \textbf{7.77}
& \textbf{65.19} & \textbf{80.84} & \textbf{20.86} \\

\bottomrule
\end{tabular}
}
\label{tab:sft_initialization}
\end{table*}

\begin{figure*}[t!]
    \centering

    \begin{minipage}{1.0\textwidth}
        \centering

        \begin{minipage}[t]{0.33\linewidth}
            \centering
            \includegraphics[width=\linewidth]
            {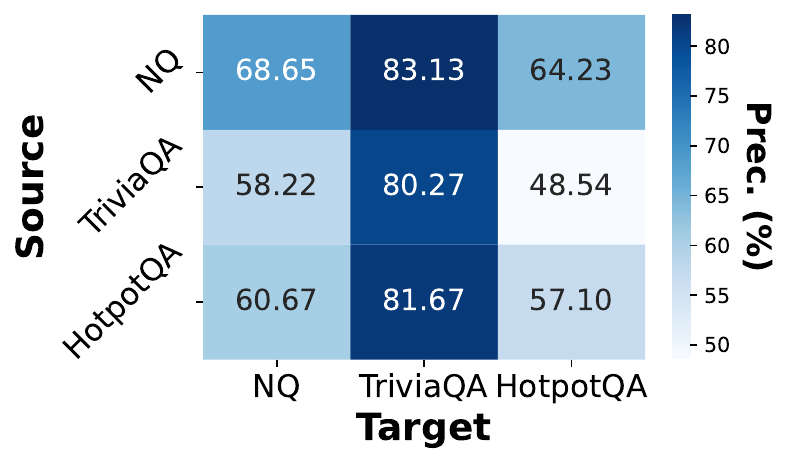}

            \small \textbf{(a) Precision}
        \end{minipage}
        \hfill
        \begin{minipage}[t]{0.33\linewidth}
            \centering
            \includegraphics[width=\linewidth]
            {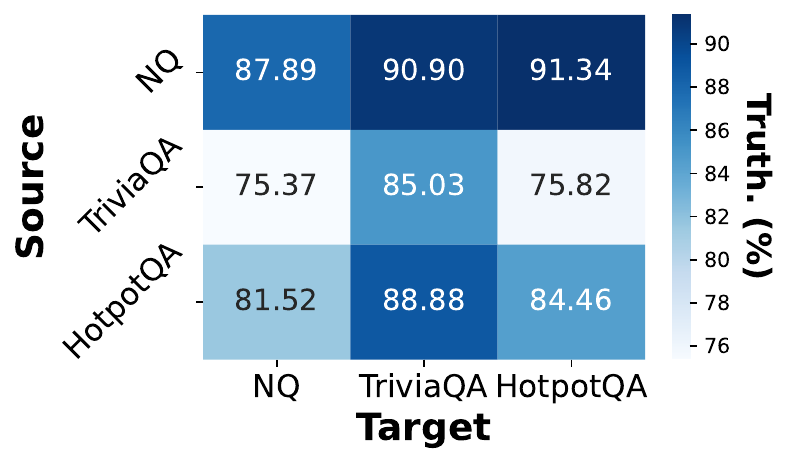}

            \small \textbf{(b) Truthfulness}
        \end{minipage}
        \hfill
        \begin{minipage}[t]{0.33\linewidth}
            \centering
            \includegraphics[width=\linewidth]
            {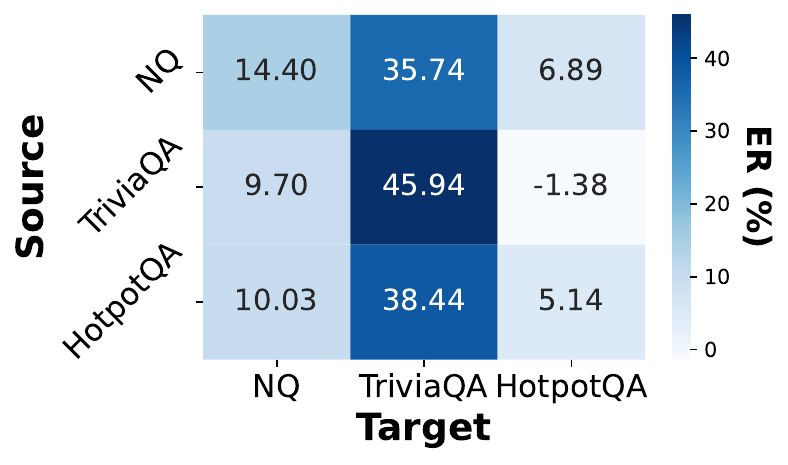}

            \small \textbf{(c) Effective Reliability}
        \end{minipage}

    \end{minipage}

    \caption{Cross-dataset generalization results of CalibraRL using LLaMA3-8B-Instruct. Rows indicate source datasets used for training, and columns indicate target datasets used for evaluation. All scores are reported as percentages, with higher values indicating better performance.}
    \label{fig:cross-dataset}
\end{figure*}

\subsection{Further Analysis}
\label{sec:ablation}
\subsubsection{Ablation Study}
As shown in Table~\ref{tab:ablation}, we evaluate the contributions of the two reward components on NQ, TriviaQA, and HotpotQA using both backbone models, keeping training data, initialization, sampling settings, and optimization hyperparameters unchanged. The \textit{w/o Answer Confidence} variant sets $\lambda_c=0$, removing the Brier-shaped penalty and reducing the answer reward to $2y_i-1$ while retaining adaptive abstention. The \textit{w/o Adaptive Abstention} variant sets $\lambda_a=0$, replacing the posterior-adaptive abstention reward with zero while retaining answer-confidence shaping. Abstaining responses remain in the sampled groups and GRPO advantage computation. Removing either component reduces average precision, truthfulness, and ER on both backbones. The full method also achieves the highest precision and truthfulness in all six model--dataset settings, although some ablated variants yield slightly higher ER on individual datasets. These results indicate that both reward components contribute to the overall performance of CalibraRL.

\subsubsection{Effect of SFT Initialization}
We investigate whether SFT initialization using the R-Tuning procedure benefits CalibraRL. We compare applying CalibraRL directly to the Instruct backbone with first performing SFT following R-Tuning and then applying CalibraRL, denoted as SFT + CalibraRL. As shown in Table~\ref{tab:sft_initialization}, directly applying CalibraRL improves average precision, truthfulness, and ER across the three benchmarks by 8.58, 11.03, and 4.93 percentage points, respectively, on LLaMA3-8B-Instruct. The corresponding gains on Qwen2.5-7B-Instruct are 3.68, 5.26, and 3.06 percentage points. These results favor applying CalibraRL directly to the Instruct backbones over using R-Tuning-based SFT initialization under our experimental setup.

\subsubsection{Confidence Calibration Analysis}
Table~\ref{tab:ece} reports ECE before and after CalibraRL training on three QA benchmarks. CalibraRL reduces ECE on all three benchmarks with both backbones. The average ECE decreases by 24.9\% for LLaMA3-8B-Instruct and 35.1\% for Qwen2.5-7B-Instruct, with an overall reduction of 30.8\% across both backbones and all three benchmarks. These results indicate improved confidence calibration on answered examples across the evaluated backbones and QA benchmarks.

\begin{table}[t]
\centering
\small
\renewcommand{\arraystretch}{1.0}
\caption{
    ECE performance before and after
    CalibraRL training on three datasets.
    Lower values indicate better calibration.
}
\label{tab:ece}
\setlength{\tabcolsep}{4pt}
\begin{tabular*}{\columnwidth}{
    @{\extracolsep{\fill}} l ccc @{}
}
\toprule
\textbf{Stage}
& \textbf{NQ}
& \textbf{TriviaQA}
& \textbf{HotpotQA} \\
\hline

\rowcolor[HTML]{F0F0F0}
\multicolumn{4}{c}{
    \mbox{\textit{Base Model: LLaMA3-8B-Instruct}}
} \\
Original Model & 0.318 & 0.147 & 0.433 \\
\textbf{Ours} & \textbf{0.193} & \textbf{0.133} & \textbf{0.348} \\
\hline

\rowcolor[HTML]{F0F0F0}
\multicolumn{4}{c}{
    \mbox{\textit{Base Model: Qwen2.5-7B-Instruct}}
} \\
Original Model & 0.444 & 0.239 & 0.513 \\
\textbf{Ours} & \textbf{0.266} & \textbf{0.172} & \textbf{0.338} \\
\bottomrule
\end{tabular*}
\end{table}

\subsubsection{Cross-Dataset Generalization}

We evaluate the generalization of CalibraRL across QA datasets using LLaMA3-8B-Instruct. Specifically, we train a separate model on the training split of each source dataset and evaluate it on the held-out evaluation splits of all three datasets without further training or parameter updates. This setting examines how the learned answering and abstention behavior transfers across data distributions. As shown in Figure~\ref{fig:cross-dataset}, diagonal entries report in-domain performance, whereas off-diagonal entries report cross-dataset performance. Models trained on NQ and HotpotQA achieve positive ER on both out-of-domain datasets. However, transfer performance varies across source--target pairs: the model trained on TriviaQA achieves an ER of 9.70 on NQ and $-1.38$ on HotpotQA. These results suggest that CalibraRL retains useful answering behavior in several cross-dataset settings, although its transfer performance depends on the source and target distributions.
 
\begin{figure}[t!]
    \centering
    \includegraphics[width=0.5\textwidth]{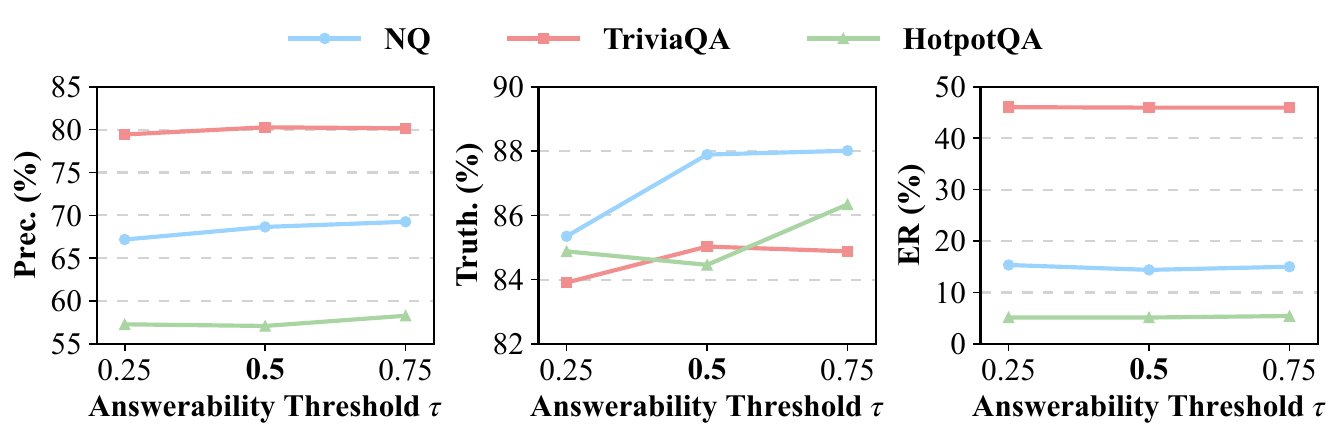}
\caption{Sensitivity of CalibraRL to the reliability threshold $\tau$ using LLaMA3-8B-Instruct. The bold tick label indicates the threshold used in the main experiments. All other hyperparameters are held fixed.}
    \label{fig:answerability-threshold}
\end{figure}

\begin{figure}[t!]
    \centering
    \includegraphics[width=0.5\textwidth]{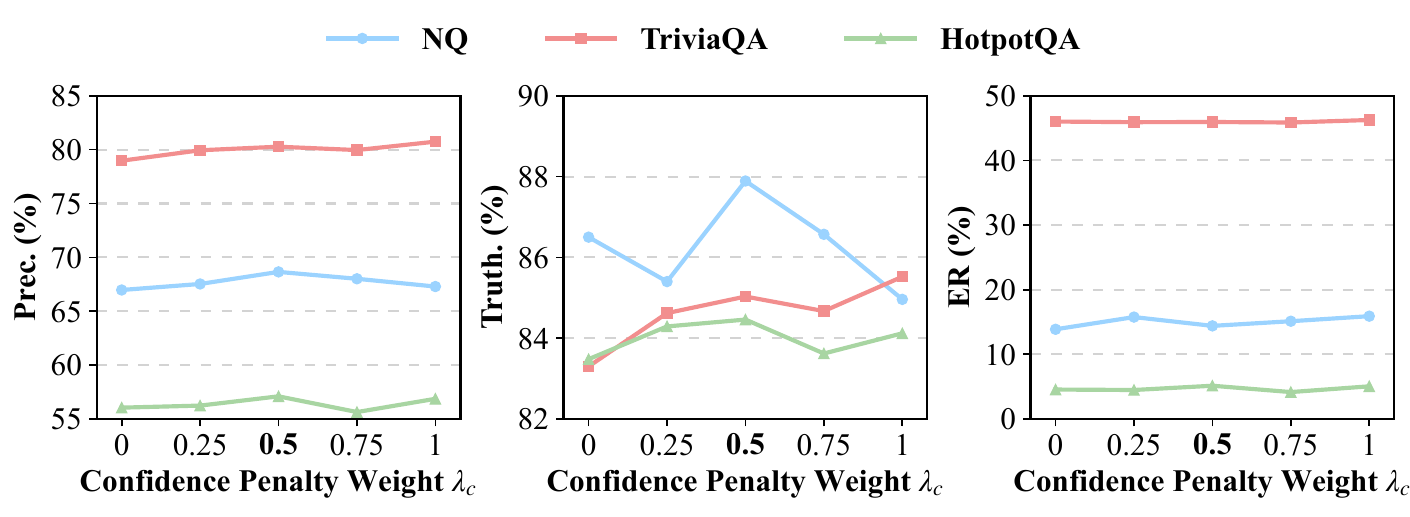}
   \caption{Effect of the confidence penalty weight $\lambda_c$ using LLaMA3-8B-Instruct. Setting $\lambda_c=0$ removes the confidence penalty while retaining the posterior-adaptive abstention reward. The bold tick label indicates the weight used in the main experiments. All other hyperparameters are held fixed.}
    \label{fig:confidence-penalty-weight}
\end{figure}

\subsubsection{Effect of the Reliability Threshold}
The reliability threshold $\tau$ in Eq.~\eqref{eq:posterior_tail} defines the cutoff used to compute the posterior probability that conditional answer accuracy falls below $\tau$. This probability is then used to compute the abstention reward in Eq.~\eqref{eq:abstention_reward}. We vary $\tau$ over $\{0.25,0.5,0.75\}$ and evaluate its effect on NQ, TriviaQA, and HotpotQA using LLaMA3-8B-Instruct. As shown in Figure~\ref{fig:answerability-threshold}, precision and truthfulness on NQ increase as $\tau$ increases, while performance on TriviaQA changes only slightly. On HotpotQA, truthfulness decreases at $\tau=0.5$ and then increases at $\tau=0.75$, while precision varies more modestly. ER remains relatively stable across the tested thresholds on all three benchmarks. These results indicate that the effect of $\tau$ varies across datasets and metrics, with no single tested threshold achieving the best performance in every setting.

\subsubsection{Effect of the Confidence Penalty Weight}
The confidence penalty weight $\lambda_c$ controls the contribution of the squared-error penalty in the answer reward,
$r_{\mathrm{ans}}(o_i,x)=2y_i-1-\lambda_c(c_i-y_i)^2$,
where $y_i\in\{0,1\}$ indicates answer correctness and $c_i$ denotes response confidence. We vary $\lambda_c$ over $\{0,0.25,0.5,0.75,1\}$ on NQ, TriviaQA, and HotpotQA using LLaMA3-8B-Instruct, with all other hyperparameters held fixed. Setting $\lambda_c=0$ recovers the \textit{w/o Answer Confidence} ablation by removing the confidence penalty while retaining the posterior-adaptive abstention reward. As shown in Figure~\ref{fig:confidence-penalty-weight}, $\lambda_c=0.5$ yields the highest precision and truthfulness on NQ, although some other tested values yield higher ER. On TriviaQA, precision and truthfulness peak at $\lambda_c=1$, while ER changes only slightly across the tested weights. Performance on HotpotQA also varies modestly. Overall, the effect of the confidence penalty depends on the dataset and metric, with no single tested weight performing best across all settings.

\begin{figure}[t]
    \centering
    \includegraphics[width=1.0\linewidth]{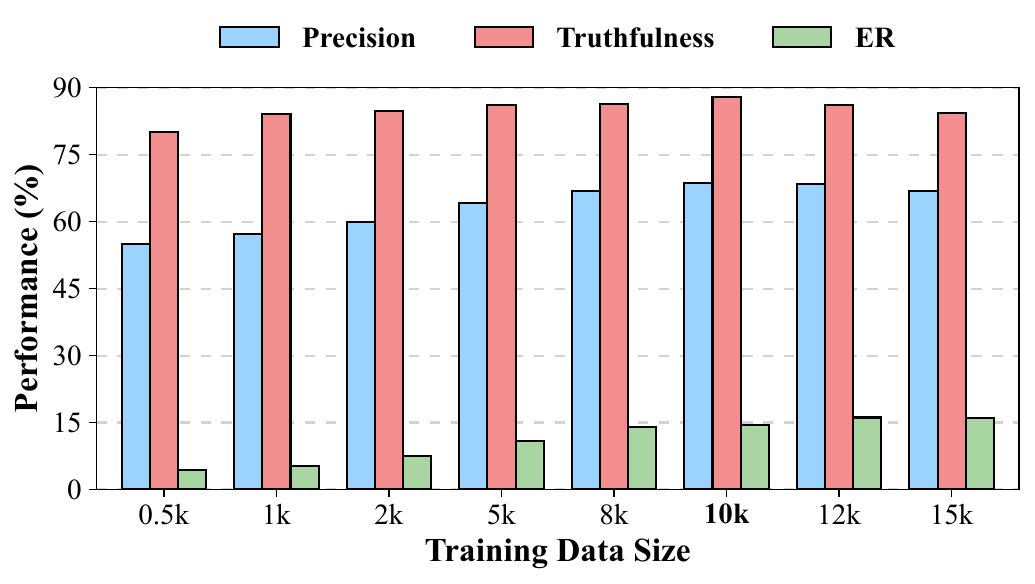}
   \caption{Effect of training data size using LLaMA3-8B-Instruct on NQ. The bold tick label indicates the training size of 10k examples used in the main experiments.}
    \label{fig:training_data_size}
\end{figure}
\subsubsection{Effect of Training Data Size}
We investigate the effect of training data size using LLaMA3-8B-Instruct on NQ by varying the number of training examples from 0.5k to 15k. As shown in Figure~\ref{fig:training_data_size}, Precision, Truthfulness, and ER generally improve as the training set grows to 10k examples. At this size, CalibraRL achieves a precision of 68.65\%, a truthfulness of 87.89\%, and an ER of 14.40\%. Increasing the training set beyond 10,000 examples further improves ER but reduces precision and truthfulness. These results indicate that larger training sets do not consistently improve all metrics. In our main experiments, we use 10k training examples, which provide strong overall performance without requiring a larger training set.

\section{Conclusion}

We identify a limitation of using fixed rewards for incorrect answers and abstentions: incorrect responses vary in confidence, while queries on which the model abstains differ in estimated answerability. We propose CalibraRL, which combines a Brier-based confidence penalty with a posterior-adaptive abstention reward derived from group-level correctness evidence. By reusing training rollouts, CalibraRL requires no additional annotations indicating whether the model knows the answer to each question. Experiments on three QA benchmarks with two instruction-tuned backbones demonstrate consistent improvements in precision and truthfulness over TruthRL. Further analysis shows improved confidence calibration on answered examples across both backbone models. These findings highlight the value of incorporating confidence and estimated answering reliability into reinforcement learning rewards for more reliable answering and abstention. Future work will explore robustness to noisy correctness feedback and extend CalibraRL to long-form generation.
\section*{Acknowledgements}
This work was funded by the National Natural Science Foundation of China (NSFC) under Grants No. 62406013 and No. U25B2070, the Beijing Advanced Innovation Center Funds for Future Blockchain and Privacy Computing, and the Fundamental Research Funds for the Central Universities.

\bibliographystyle{ACM-Reference-Format}
\bibliography{sample-base}

\appendix
\newpage

\section{Algorithm Details}
\label{app:algorithm}
Algorithm~\ref{alg:calibrarl} summarizes CalibraRL training.
Responses receive confidence-aware answer rewards or
posterior-adaptive abstention rewards, with the posterior
derived from a fixed prior and the correctness outcomes
of the group's non-abstaining responses.

\begin{algorithm}[h]
\caption{CalibraRL}
\label{alg:calibrarl}
\small
\begin{algorithmic}[1]
\Require Training data $\mathcal{D}$, initial policy $\pi_\theta$,
fixed reference policy $\pi_{\mathrm{ref}}$,
group size $G$, prior parameters $\alpha_0,\beta_0$,
reliability threshold $\tau$, reward weights
$\lambda_c,\lambda_a$, clipping parameter $\epsilon$,
KL weight $\beta_{\mathrm{KL}}$, stability constant
$\varepsilon_{\mathrm{std}}$
\Ensure Trained policy $\pi_\theta$

\For{each training iteration}
    \State $\theta_{\mathrm{old}} \gets \theta$
    \State Sample a training batch $\mathcal{B}$ from $\mathcal{D}$
    \For{each $(x,\mathcal{A}) \in \mathcal{B}$}
        \State Sample $\mathcal{G}_x=\{o_i\}_{i=1}^{G}$ from $\pi_{\theta_{\mathrm{old}}}(\cdot\mid x)$
        \State Identify abstentions $a_i \gets a(o_i)$ for all $i$
        \State $\mathcal{I}_x \gets \{i:a_i=0\}$
        \For{each $i \in \mathcal{I}_x$}
            \State $y_i \gets y(o_i,\mathcal{A})$
            \State $c_i \gets c_{\theta_{\mathrm{old}}}(x,o_i)$ using Eq.~\eqref{eq:response_confidence}
            \State $r_i \gets 2y_i-1-\lambda_c(c_i-y_i)^2$
        \EndFor
        \State $C_x \gets \sum_{i\in\mathcal{I}_x}y_i$
        \State $W_x \gets \sum_{i\in\mathcal{I}_x}(1-y_i)$
        \State $q_x \gets I_\tau(\alpha_0+C_x,\beta_0+W_x)$
        \For{each $i$ such that $a_i=1$}
            \State $r_i \gets \lambda_a(2q_x-1)$
        \EndFor
        \State $\bar{r} \gets G^{-1}\sum_{i=1}^{G}r_i$
        \State $s_r \gets \operatorname{Std}(r_1,\ldots,r_G)$
        \For{$i=1,\ldots,G$}
            \State $\mathbb{A}_i \gets (r_i-\bar{r})/(s_r+\varepsilon_{\mathrm{std}})$
        \EndFor
        \State Store $\mathcal{G}_x$ and its fixed advantages $\{\mathbb{A}_i\}_{i=1}^{G}$
    \EndFor

    \For{each policy update on the sampled batch}
        \State Compute token-level ratios $\rho_{i,t}(\theta)$ using Eq.~\eqref{eq:importance_ratio}
        \State Compute clipped terms $\ell_{i,t}(\theta)$ using Eq.~\eqref{eq:clipped_surrogate}
        \State Compute $\mathcal{L}_{\mathrm{pg}}(\theta)$ over the batch using Eq.~\eqref{eq:grpo_policy_loss}
        \State Compute $\mathcal{L}_{\mathrm{KL}}(\theta)$ against $\pi_{\mathrm{ref}}$ using Eq.~\eqref{eq:grpo_kl_loss}
        \State $\mathcal{L}_{\mathrm{train}} \gets \mathcal{L}_{\mathrm{pg}}+\beta_{\mathrm{KL}}\mathcal{L}_{\mathrm{KL}}$
        \State Update $\theta$ by minimizing $\mathcal{L}_{\mathrm{train}}$, keeping rewards and advantages fixed
    \EndFor
\EndFor
\State \Return $\pi_\theta$
\end{algorithmic}
\end{algorithm}

\section{More Experimental Settings}

\subsection{Datasets}
\label{app:datasets}
We conduct experiments on three widely used QA datasets that cover both single-hop and multi-hop question-answering scenarios. Table~\ref{tab:qa-datasets} summarizes the key statistics of these datasets. Specifically, \textbf{NQ}~\cite{nq} and \textbf{TriviaQA}~\cite{tqa} are representative single-hop datasets, where each question can typically be answered using information from a single passage retrieved from the corpus. These datasets primarily evaluate a model’s ability to locate and extract factual evidence efficiently. In contrast, \textbf{HotpotQA}~\cite{hotpotqa} is a challenging multi-hop dataset that requires integrating and reasoning over multiple pieces of evidence distributed across different documents to derive the final answer. This dataset is particularly useful for testing a model’s reasoning and compositional understanding capabilities. 
Together, these datasets enable us to evaluate the
answering reliability and abstention behavior of
CalibraRL across factual and multi-hop QA settings.
\begin{table}[t]
\small
\centering
\caption{Dataset statistics, including reasoning types and the number of examples in each split.}
\begin{tabular}{lcccc}
\toprule
\textbf{Dataset} & \textbf{Type}&\textbf{\# Train} & \textbf{\# Dev} & \textbf{\# Test} \\
\midrule
NQ      &  single-hop & 79.1k  & 8.7k  & 3.6k  \\
TriviaQA &  single-hop& 78.7k  & 11.3k  & 8.8k \\
HotpotQA  & multi-hop& 88.9k  & 5.6k  & 5.6k  \\
\bottomrule
\end{tabular}
\label{tab:qa-datasets}
\end{table}

\subsection{Evaluation Metrics}
\label{app:evaluation_metrics}

Let $N$ denote the total number of evaluation examples, and let $N_{\mathrm{cor}}$, $N_{\mathrm{inc}}$, and $N_{\mathrm{abs}}$ denote the numbers of correct answers, incorrect answers, and abstentions, respectively. Thus, $N=N_{\mathrm{cor}}+N_{\mathrm{inc}}+N_{\mathrm{abs}}$, and the number of answered examples is $N_{\mathrm{ans}}=N_{\mathrm{cor}}+N_{\mathrm{inc}}$.

\paragraph{\textbf{Precision}}
Precision measures the proportion of correct answers among answered examples. It evaluates answer accuracy conditional on the model choosing to answer.

\paragraph{\textbf{Truthfulness}}
Truthfulness measures the proportion of evaluation examples on which the model either answers correctly or abstains: $\mathrm{Truthfulness}=
    \frac{N_{\mathrm{cor}}+N_{\mathrm{abs}}}{N}
    =
    1-\frac{N_{\mathrm{inc}}}{N}.$
Because abstentions contribute positively to this metric, we interpret it jointly with precision and ER.

\paragraph{\textbf{ER}}
Effective reliability assigns a utility of $+1$ to correct answers, $-1$ to incorrect answers, and $0$ to abstentions: $\mathrm{ER}
    =
    \frac{N_{\mathrm{cor}}-N_{\mathrm{inc}}}{N}$. It measures the net utility of answering across the full evaluation set.

\paragraph{\textbf{ECE}}
We compute ECE~\cite{guo2017calibration} only on answered examples. For each answered example $i$, let $c_i\in[0,1]$ denote the response confidence and $y_i\in\{0,1\}$ indicate whether the answer is correct. We divide the confidence interval $[0,1]$ into $M$ equal-width bins and assign each answered example to the corresponding bin. Let $B_m$ denote the set of examples in bin $m$. For each nonempty bin, empirical accuracy and average confidence are $\operatorname{Acc}(B_m)
    =
    \frac{1}{|B_m|}
    \sum_{i\in B_m} y_i$ and $\operatorname{Conf}(B_m)
    =
    \frac{1}{|B_m|}
    \sum_{i\in B_m} c_i.$
ECE is the weighted average of the absolute differences between bin accuracy and bin confidence:
\begin{equation}
    \mathrm{ECE}
    =
    \sum_{\substack{1\leq m\leq M\\|B_m|>0}}
    \frac{|B_m|}{N_{\mathrm{ans}}}
    \left|
    \operatorname{Acc}(B_m)
    -
    \operatorname{Conf}(B_m)
    \right|.
    \label{eq:metric_ece}
\end{equation}
We use $M=10$ in all experiments. Lower ECE indicates closer agreement between confidence and empirical accuracy on answered examples.





\subsection{Training Details}
\label{app:training}
We use LLaMA3-8B-Instruct~\footnote{\url{https://huggingface.co/meta-llama/Meta-Llama-3-8B-Instruct}}~\cite{grattafiori2024llama} and Qwen2.5-7B-Instruct~\footnote{\url{https://huggingface.co/Qwen/Qwen2.5-7B-Instruct}}~\cite{qwen2025qwen25technicalreport} as the backbone models following prior work~\cite{cao2025autohall,wu2026step,zhang2026redippo,xi2026confidence}. These models and their variants have been widely adopted across various tasks and applications~\cite{xiao2026staying}. We train CalibraRL using GRPO with LoRA~\cite{hu2022lora} for parameter-efficient fine-tuning. For both backbones, we set the dataset size parameter to 10k and the validation ratio to 0.1. LoRA is configured with a rank of 64, a scaling parameter of 128, and a dropout rate of 0.. We train both models for one epoch with a learning rate of $1\times10^{-6}$, a batch size of 4, and 8 gradient accumulation steps. The evaluation batch size is 32, and the warmup ratio is 0.03. The maximum prompt and response lengths are 2,048 and 16 tokens, respectively. During training, we sample 32 responses per query for LLaMA3-8B-Instruct and Qwen2.5-7B-Instruct, with sampling temperatures of 1.0 and 1.2, respectively. Both models use nucleus sampling with $p=0.95$. We set the GRPO clipping parameter to 0.2 and the KL regularization coefficient to 0.01. The confidence penalty weight is 0.5. For the posterior-adaptive abstention reward, we use a uniform Beta prior with $\alpha_0=\beta_0=1$, a reliability threshold of $\tau=0.5$, and a reward scale of $\lambda_a=0.5$. No additional confidence calibration transformation is fitted during training. We fix the random seed to 42 for all runs.

\begin{figure*}[t!]
    \centering
    \includegraphics[width=1.0\textwidth]{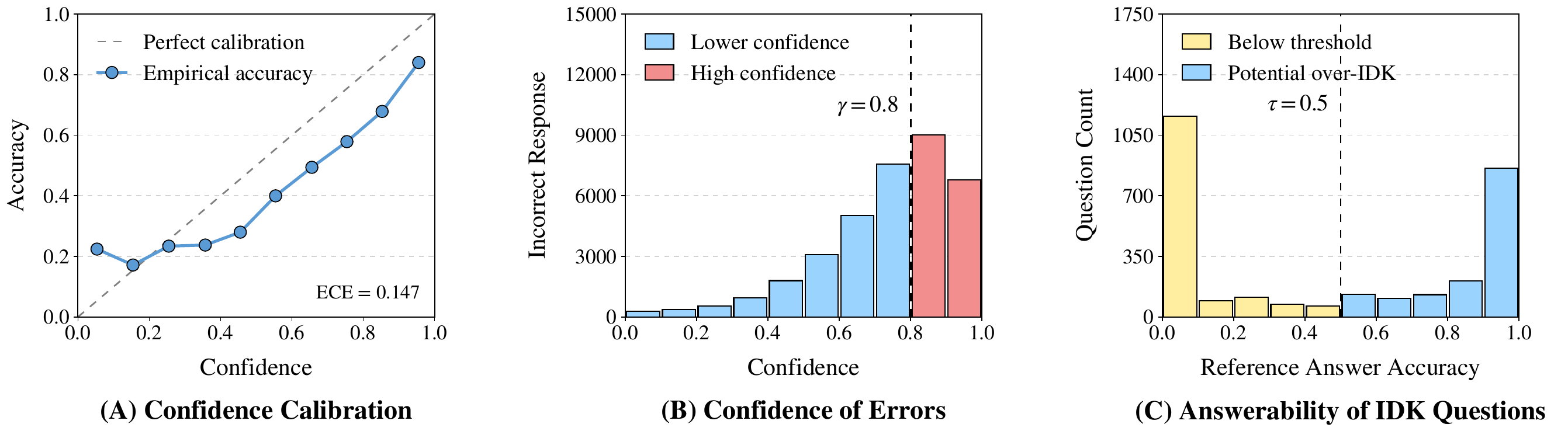}
  \caption{Empirical motivation using LLaMA3-8B-Instruct on TriviaQA. \textbf{(A)} Empirical accuracy generally increases with confidence. \textbf{(B)} Incorrect responses exhibit heterogeneous confidence. \textbf{(C)} Questions eliciting abstention differ in the accuracy of non-abstaining responses obtained through independent reference sampling.}
    \label{motivation_tqa}
\end{figure*}

\begin{figure*}[t!]
    \centering
    \includegraphics[width=1.0\textwidth]{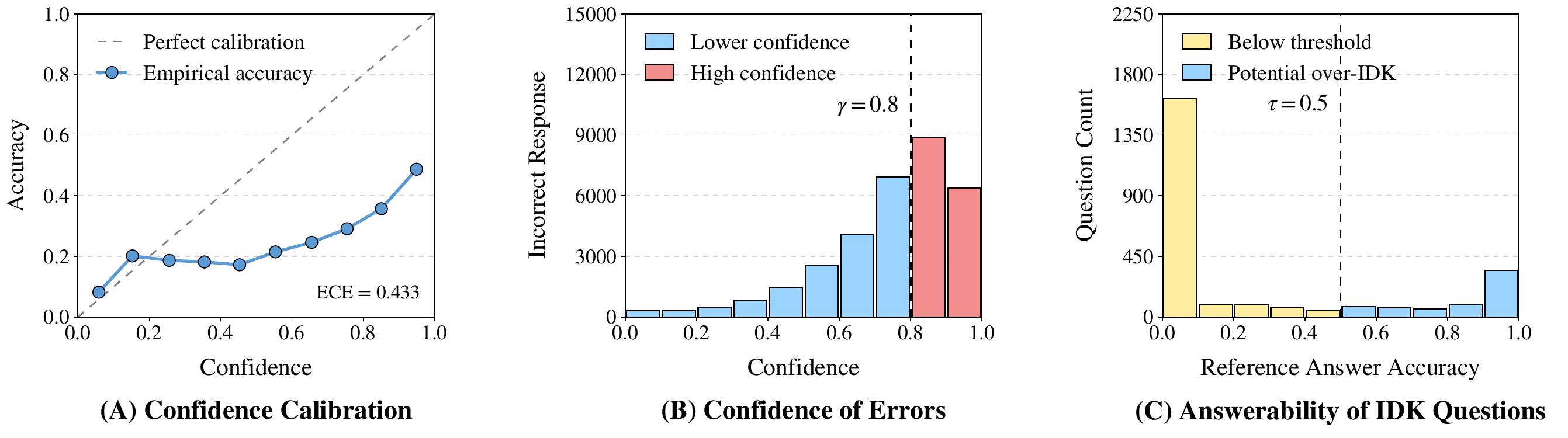}
   \caption{Empirical motivation using LLaMA3-8B-Instruct on HotpotQA. \textbf{(A)} Empirical accuracy generally increases with confidence. \textbf{(B)} Incorrect responses exhibit heterogeneous confidence. \textbf{(C)} Questions eliciting abstention differ in the accuracy of non-abstaining responses obtained through independent reference sampling.}
    \label{motivation_hotpotqa}
\end{figure*}

\subsection{Prompt}
\label{app:prompt}

\begin{table}
\caption{Prompt for the base models. }
\begin{tcolorbox}[
    colback=black!1!white,
    colframe=black!57!white,
    title=Prompt for the Base Models,
    fontupper={\fontsize{10pt}{12pt}\selectfont},
fonttitle={\fontsize{11pt}{13pt}\selectfont\bfseries}
]
\verb|<system>|\\
You are a helpful, respectful, and honest assistant. Answer the question with a couple of words using the provided documents only when you are confident the answer is correct. Otherwise, respond exactly with 'I don't know'.\\
\verb|</system>|
\\
\verb|<user>|\\
Question: \verb|{query}|\\
Documents: \\
Doc1: \verb|{Document 1}|  \\
Doc2: \verb|{Document 2}|  \\
...... \\
\verb|</user>|\\
\end{tcolorbox}
\label{tab:prompt}
\end{table}

Table~\ref{tab:prompt} presents the prompt template used for the base models. The system instruction encourages concise answers grounded in the supplied documents and requires the model to respond exactly with \emph{``I don't know''} when uncertain. The user message contains the question and the retrieved documents, which provide external evidence for answering.

\section{More Experimental Results}

\subsection{Additional Empirical Motivation}
We extend the empirical analyses in the preliminaries to TriviaQA and HotpotQA. Figure~\ref{motivation_tqa} and Figure~\ref{motivation_hotpotqa} show the confidence--accuracy relationship, the confidence distribution of incorrect responses, and reference answer accuracy for questions eliciting abstention, following the same protocol used for NQ. 

\begin{figure}[th!]
    \centering
    \includegraphics[width=\linewidth]
    {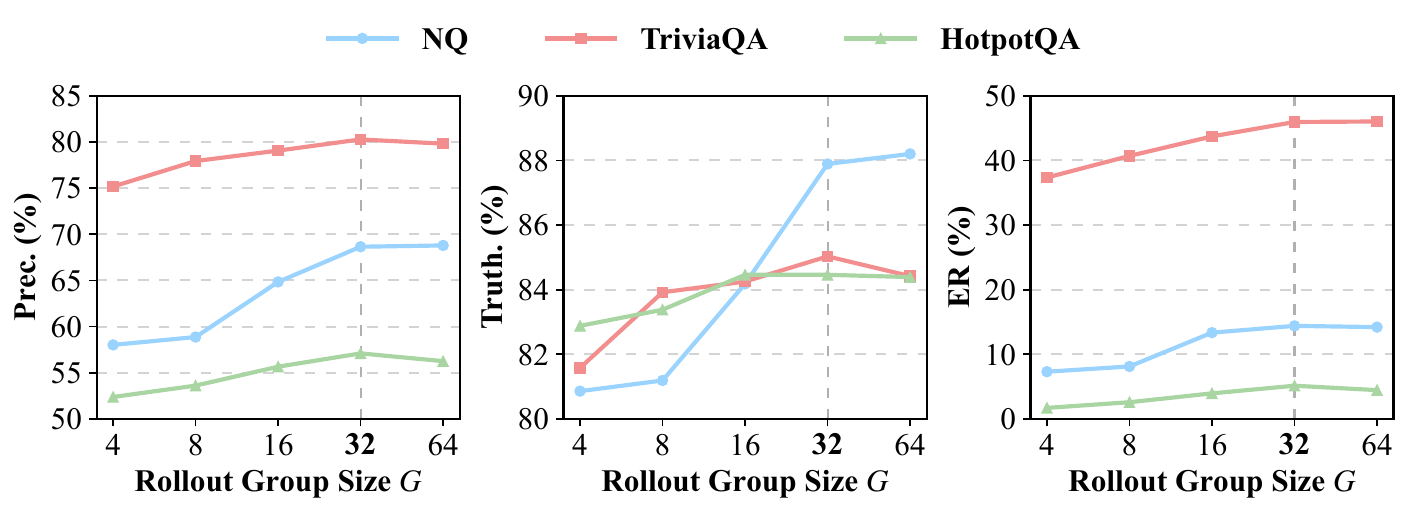}
    \caption{Effect of rollout group size $G$ on
    (a) precision, (b) truthfulness, and (c) effective
    reliability (ER) across NQ, TriviaQA, and HotpotQA.
    The vertical dashed lines indicate the default
    setting $G=32$.}
    \label{fig:group_size}
\end{figure}

\subsection{Effect of Rollout Group Size}
\label{sec:group_size}
We vary the rollout group size $G$ over $\{4,8,16,32,64\}$. Group size affects abstention reward estimation and group-relative advantage computation. As shown in Figure~\ref{fig:group_size}, increasing $G$ from $4$ to $32$ generally improves performance across datasets. Precision increases by $10.63$, $5.10$, and $4.72$ percentage points on NQ, TriviaQA, and HotpotQA, respectively, with improvements in truthfulness and ER. These results suggest that larger groups provide more informative samples for abstention reward estimation and group-relative advantage computation. Increasing $G$ to $64$ yields limited and inconsistent gains: NQ shows marginal improvements in precision and truthfulness but a slight decline in ER, TriviaQA exhibits a small increase in ER despite decreases in the other two metrics, and HotpotQA shows declines in all three metrics. Overall, $G=32$ achieves the best or near-best performance, supporting its use as the default group size.
\end{document}